\documentclass[a4paper, 10pt, conference]{ieeeconf}      

\IEEEoverridecommandlockouts                              

\usepackage{graphics} 
\usepackage{hyperref}
\hypersetup{hidelinks}
\usepackage{epsfig} 
\usepackage{amsmath} 
\usepackage{amssymb}  

\title{\LARGE \bf
Investigating the Generalizability of Assistive Robots Models over Various Tasks
}

\author{Hamid Osooli, Christopher Coco, Johnathan Spanos, Amin Majdi and Reza Azadeh
\thanks{Authors are with the Persistent Autonomy and Robot Learning (PeARL) Lab, University of Massachusetts Lowell, Lowell, MA 01854, USA
        {\tt\small \{hamid\_osooli, reza\_azadeh\}@uml.edu, \{christopher\_coco, johnathan\_spanos, amin\_majdi\}@student.uml.edu}}%
}

\begin{document}
\bstctlcite{IEEEexample:BSTcontrol}

\maketitle
\thispagestyle{empty}
\pagestyle{empty}

\begin{abstract}

In the domain of assistive robotics, the significance of effective modeling is well acknowledged. Prior research has primarily focused on enhancing model accuracy or involved the collection of extensive, often impractical amounts of data. While improving individual model accuracy is beneficial, it necessitates constant remodeling for each new task and user interaction. In this paper, we investigate the generalizability of different modeling methods. We focus on constructing the dynamic model of an assistive exoskeleton using six data-driven regression algorithms. Six tasks are considered in our experiments, including horizontal, vertical, diagonal from left leg to the right eye and the opposite, as well as eating and pushing. We constructed thirty-six unique models applying different regression methods to data gathered from each task. Each trained model's performance was evaluated in a cross-validation scenario, utilizing five folds for each dataset. These trained models are then tested on the other tasks that the model is not trained with. Finally the models in our study are assessed in terms of generalizability. Results show the superior generalizability of the task model performed along the horizontal plane, and decision tree based algorithms.

\end{abstract}

\section{Introduction}

The importance of modeling in the realm of assistive robotics is a well-recognized aspect that is crucial for effective control and user interaction. Assistive robots, particularly those in constant interaction with human users, present unique challenges in modeling. Unlike conventional rigid body systems governed by standard physics, these robots involve complex interactions between the user and the robot. This complexity necessitates the adoption of data-driven modeling techniques, which are frequently used in robotics research~\cite{wu2012semi, riedel2019comparing, kwiatkowski2019task, zhang2020sufficiently}.

One of the primary challenges in these systems is the variability of the user actions. For a model to be task generalizable, it ideally needs exposure to a comprehensive range of movement trajectories. Previous studies, such as \cite{kwiatkowski2019task}, demonstrated the need for an extensive set of motion trajectories, by training robots on a vast array of random movement trajectories, enhancing their ability to perform unforeseen tasks. However, this approach is impractical in real-world scenarios, especially when considering the vast data requirements and the unique challenges posed by assistive robots used by individuals with disabilities.

To enhance the accuracy of the trained model, Zhang et al.~\cite{zhang2020sufficiently} proposed an optimization problem, where model accuracy is a constraint. While novel methods to increase accuracy are valuable, they do not inherently lead to a generalizable model that minimizes the need for additional data collection. This highlights the need for an in-depth study into generalizability that provides sufficient data for developing a generalizable model.

\begin{figure}[t]
    \centering
\includegraphics[width=.7\columnwidth]{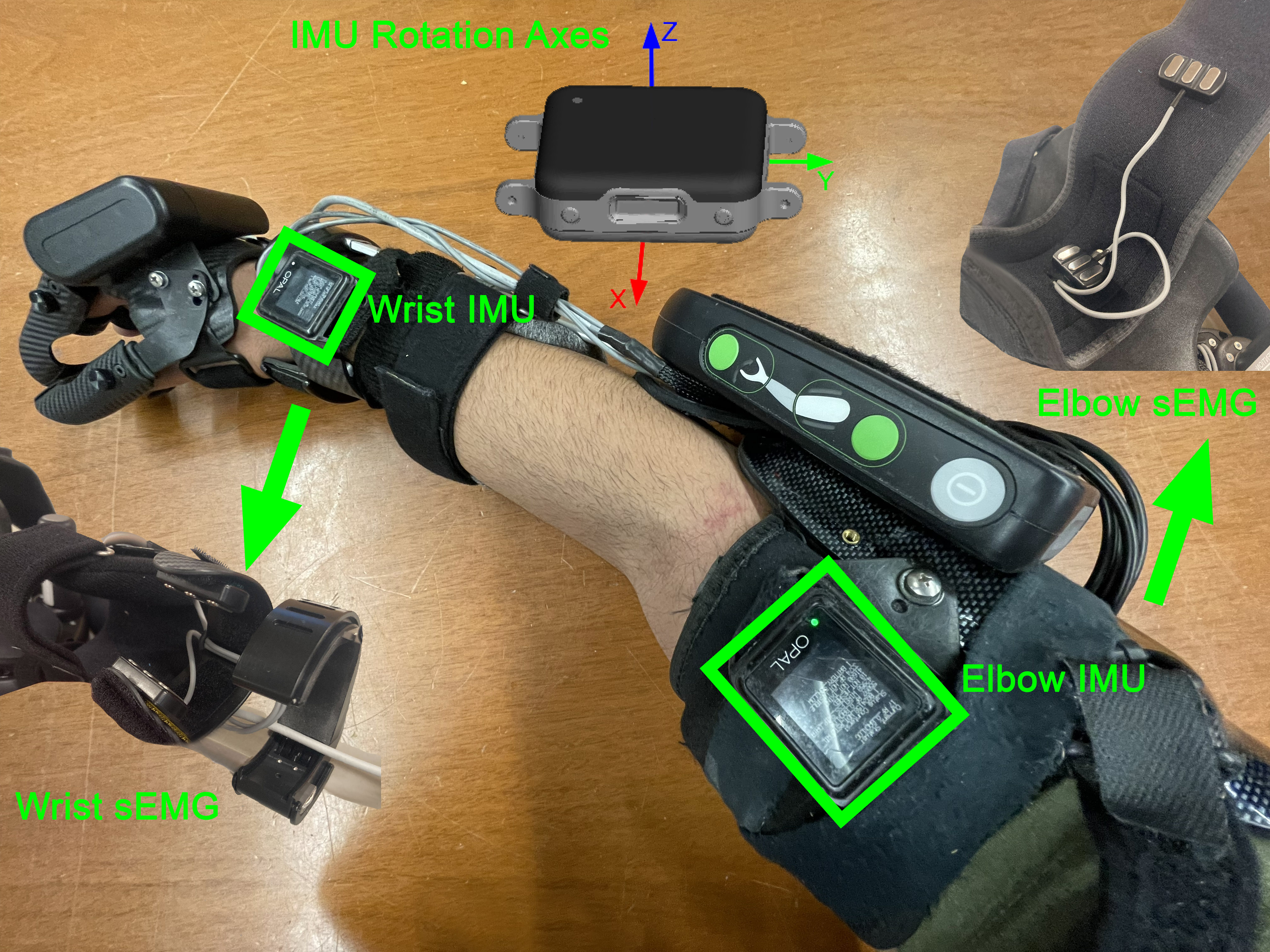}
    \caption{\small{Myopro prosthetic augmented with IMU sensors for data collection. Figure annotation highlights the placement of sEMG sensors and the rotational axes of APDM Opal IMU sensors.}}
    \label{fig:exoimu}
\end{figure}

This paper introduces a comprehensive study on the generalizability of the tasks models. We focus on upper-limb exoskeletons, specifically using the MyoPro 2 Motion-G, a 2 DOF exoskeleton, equipped with two APDM opal IMU sensors. Our study encompasses six algorithms namely Locally Weighted Projection Regression~(LWPR), K-Nearest Neighbours~(KNN), Support Vector Regression~(SVR), eXtreme Gradient Boosting~(XGBoost), Multi Layer Perceptron~(MLP), and Gaussian Process Regression~(GPR). The performance of each algorithm is evaluated on six different tasks (Horizontal~(H), Vertical~(V), diagonal from Left leg to Right eye~(LR), diagonal from Right leg to Left eye~(RL), Eating~(E), and Pushing~(P)). We use the R-squared score to assess the effectiveness and generalizability of each task model and algorithm.

Results show that the task models performed along the horizontal plane, and decision tree based algorithms are superior in terms of the generalizability. These findings are practical for developing strategies that can enhance the effectiveness and adaptability of models across diverse scenarios.

\section{Related Work}
The generalizability of the machine learning models helps on reducing the need for repetitive training and data collection. This concept has been addressed in other fields that study the generalizability of deep learning models in visual tasks~\cite{alhamoud2022generalizability}. However, generalizability remains a largely unexplored factor in modeling the interaction between the users and the prosthetic robots.

Siu et al.,~\cite{siu2018implementation} introduced a non-adaptive controller that integrates pressure and sEMG data gathered during training to construct a K-nearest neighbors (KNN) classifier. This classifier was built on fourteen signal features derived from each of the six sEMG sensors. The sEMG features were normalized through mean subtraction and division by standard deviation. 
Additionally, an adaptive controller was employed to update the sEMG mapping using the KNN classifier. The effectiveness of the proposed model was assessed through a tabletop book-shelving task. The controller exhibited adaptability to user-specific physiological changes, such as fatigue. The authors then proposed a Learning from Demonstration (LfD)~\cite{hertel2022robot,ravichandar2019skill} approach in which the user demonstrates the task for the robot, enabling it to learn and be trained. Although practical in many robotic experiments, LfD approach may be less effective, especially in cases where the prosthesis is being used for an impaired limb.

\begin{figure}[h!]
    \centering
    \includegraphics[width=\columnwidth]{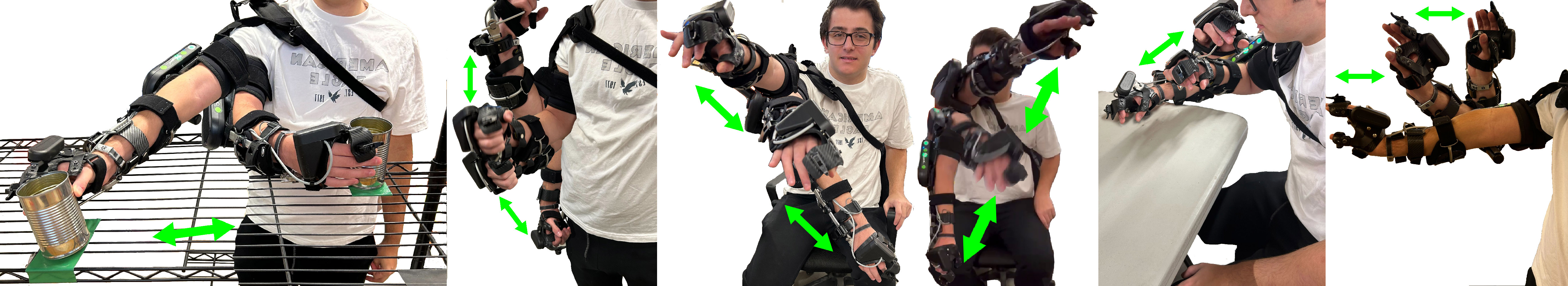}
    \caption{\small{Overview of the diverse tasks employed for data acquisition from test subjects, including Horizontal (H), Vertical (V), diagonal from Left leg to Right eye (LR), diagonal from Right leg to Left eye (RL), Eating (E), and Pushing (P).}}
    \label{fig:tasks}
\end{figure}

To achieve task generalizable capability, the model necessitates exposure to a comprehensive set of diverse trajectories. Kwiatkowski and Lipson~\cite{kwiatkowski2019task} addressed this requirement by training the robot on a dataset comprising 1000 randomly generated trajectories, enhancing its adaptability to unforeseen tasks. Zhang et al.~\cite{zhang2020sufficiently} approached the modeling task as an optimization problem, treating model accuracy as a constraint. Since pre-covering the complete state-space with data is impractical~\cite{nguyen2011model}, many of the works in this area proposed online model learning~\cite{cao2006neural,nguyen2009model,reinhart2009reaching}, as a way to gather more data for modeling.

While collection of huge amounts of motion trajectory data is feasible in various scenarios, it poses a challenge for assistive robots, particularly exoskeletons. These devices often help users with limited mobility, such as individuals with disabilities, who may not engage in extensive movements suitable for data collection. Therefore, the strategic design of efficient tasks that results in sufficient amount of data for training a generalized model becomes important in such contexts.

\section{Problem Formulation}\label{probForm}

A comprehensive model capturing the interaction between the user and the exoskeleton necessitates an encompassing description of the device states across its elbow and wrist degrees of freedom. The system’s formulation, encapsulating robot states, inputs, and user inputs can be formulated as:
\begin{equation}
x_{t+1}= x_t + f(x_t, u_t, v_t) + \eta_t,\;\;\;\eta_t\sim\mathcal{N}(0, \Sigma_\eta),
    \label{eq:prob_form}
\end{equation}

\noindent where $f$ defines the unknown dynamic evolution of the interaction between the user and the exoskeleton over time $t$, $x \in \mathbb{R}^{L}$ is the state vector (elbow/wrist angles and angular velocities), and $v \in \mathbb{R}^{N}$ is the user's action incorporating biceps/triceps sEMG measurements for the elbow and  the wrist, $u \in \mathbb{R}^{M}$ is the robots action vector including the thresholds considered on the difference between the sEMG signals of the biceps and triceps or opening and closing of the hand. This sEMG threshold triggers the robot's assistance when surpassed. The uncertainties from IMUs are modeled by an additive white Gaussian noise~\cite{nirmal2016noise}. Potential noises from the exoskeleton, however, were compensated by the built-in mechanisms, allowing for their exclusion from our noise consideration framework.

\section{Dynamic Model Learning via Regression}\label{sec:Dynamic Model Learning through Regression}
Modeling the dynamics of exoskeleton robots is more challenging due to the presence of human actions and the interaction between the device and the human. To accurately model $f$, in this paper, we utilize six different regression methods, namely Locally Weighted Projection Regression (LWPR), K-Nearest Neighbours (KNN), Support Vector Regression (SVR), eXtreme Gradient Boosting (XGBoost), Multi Layer Perceptron (MLP), and Gaussian Process Regression (GPR). 

Each model incorporates an input vector made of the states $x_t$, robot's actions $u_t$ and the user's actions $v_t$, represented as the vector $\Tilde{x}_t = [x_t, u_t, v_t] \in \mathbb{R}^{L+M+N}$. For the training targets, we use the difference between the current and future state vectors: $\Delta{x_t} = x_{t+1} - x_t \in \mathbb{R}^L$. The target dataset is defined as $T_{1:t}^L = \{f(\Tilde{x}_1), \dots, f(\Tilde{x}_t)\}$, and a new input point at which the model is queried is shown with $\Tilde{x}_*$.

\subsection{Locally Weighted Projection Regression (LWPR)}
LWPR~\cite{vijayakumar2005lwpr} is an algorithm that achieves nonlinear function approximation in high dimensional spaces with redundant and irrelevant input dimensions having little to no impact on the output. It considers $R$ locally linear models for approximation of the function as: $f = \mathbb{E}\{\bar{f}_k|\Tilde{x}_*\} = \sum_{k=1}^R \bar{f}_k p(k|\Tilde{x}_*)$. From the Bayes' theorem, when we query a new input point $\Tilde{x}_*$, the probability of the model $k$ can be expressed as:
\begin{equation}
    p(k|\Tilde{x}_*) = \frac{p(k|\Tilde{x}_*)}{p(\Tilde{x}_*)} = \frac{p(k|\Tilde{x}_*)}{\sum_{k=1}^R p(k|\Tilde{x}_*)} = \frac{w_k}{\sum_{k=1}^R w_k}.
\end{equation}

\noindent Hence, 

\begin{equation}
f(\Tilde{x}_*) \sim \frac{\sum_{k=1}^{R}w_k\bar{f}_k(\Tilde{x}_*)}{\sum_{k=1}^{R}w_k},
    \label{eq:lwprNorm}
\end{equation}

\noindent we have $\bar{f}_k = \bar{x}_k^\top \hat{\theta}_k$ and $\Bar{x}_k = [(\Tilde{x}_* - c_k)^\top, 1]^\top$, in which $\hat{\theta}_K$ consists of the estimated parameters of the model, and $c_k$ is the center of the $k$-th linear model. $w_k$ is the weight that determines whether a data point $\Tilde{x}_*$ is into the region of validity of the model $k$. This is similar to a receptive field, and is defined by a Gaussian kernel

\begin{equation}
    w_k = \exp(-\frac{1}{2}(\Tilde{x}_* - c_k)^\top D_k(\Tilde{x}_* - c_k)),
\end{equation}

\noindent where $D_k$ is the distance matrix and should be positive definite. In the learning procedure, the shape of the $D_k$ and the $\hat{\theta}_k$ parameters of the local models are adjusted to minimize the error between the predicted values and the observed targets~\cite{nguyen2009model}. We initialize $D_k$ as $D_k = rI_{L+M+N}$, where $r$ is a constant value tuned based on the model performance and $I$ is the identity matrix with the same size as the inputs.

\subsection{K-Nearest Neighbours (KNN)}

The KNN algorithm~\cite{cover1967nearest} is based on the distance-weighted nearest neighbor estimation, where k most similar values of the input data~$\Tilde{x}$ are used for the prediction of the diameter distribution of $f(\Tilde{x})$. The similarity of the data is measured by their distance as

\begin{equation}
    d_{ij} = \sum_{l=1}^L c_l \lVert(\Tilde{x}_{il} - \Tilde{x}_{jl})\rVert,
\end{equation}

\noindent where $\Tilde{x}_l$ is the input, and $c_l$ is the coefficient for the input. Then the distances $d$ are sorted based on the weight calculated by

\begin{equation}
    w_{ij} = \frac{(\frac{1}{1+d_{ij}})^p}{\sum_{i=1}^k(\frac{1}{1+d_{ij}})^p}, \forall \; i \neq j,
\end{equation}
\noindent where $k$ is the number of the nearest neighbors used, and $p$ is the weighting parameter of distance. The weighting parameter $p$ determines how fast should the weights of the nearest neighbors decrease when the distance $d_{ij}$ increases, and weights should sum to one.

\subsection{Support Vector Regression (SVR)}
SVR~\cite{awad2015support} is an algorithm that belongs to the family of support vector machines. In SVR, the regression function $f(\Tilde{x})$ is estimated by the hyper plane $h$:
\begin{equation}
    f(\Tilde{x}) = h^\top \Tilde{x}_* + b \text{ with } h\in \mathbb{R}^{L+M+N}, b\in\mathbb{R}.
\end{equation}

Exploiting the structural risk minimization~\cite{vapnik1999nature}, the generalization accuracy of the SVR is optimized on the empirical error and flatness of the $f(\Tilde{x})$ which is the result of the small values for $h$. Therefore the SVR aims to include the dataset patterns inside an $\epsilon$-tube while minimizing the $\lVert h \rVert^2$. We can formulate this as an optimization problem:
\begin{align}
    &{\text{minimize}}\frac{1}{2}\lVert h \rVert^2 + C\sum_{i=1}^l (\xi_i + \xi_i^*),\nonumber\\&     
     \text{s.t. }f(\Tilde{x}_i) - h^\top \Tilde{x}_i - b \leq \epsilon + \xi_i,\nonumber\\&
     h^\top \Tilde{x}_i + b - f(\Tilde{x}_i)\leq \epsilon + \xi_i^*,\nonumber\\&
     \xi_i, \xi_i^*\geq 0,\;\;\;\;i=1, \dots, l,
\end{align}
\noindent where $C, \epsilon,$ and $\xi, \xi^*$ are the cost for the trade-off between the empirical error and the flatness of the $f(\Tilde{x})$, the $\epsilon$-tube size, and slack variables. Adding the Lagrangian multipliers $\alpha$ and $\alpha^*$ transforms the quadratic programming problem into a dual optimization. Furthermore, SVR is capable of non-linear function approximation by employing a kernel function $k_{f}(\Tilde{x}_i, \Tilde{x}_j)$. Thus the SVR estimates $f$ as:
\begin{equation}
    f(\Tilde{x}) \sim \sum_{i=1}^s (\alpha - \alpha^*)k_{f}(\Tilde{x}_i, \Tilde{x}_j) +b,
\end{equation}

\noindent where $s$ is the number of the support vectors, and $b$ is a constant~\cite{kim2008response}. In this paper we use a composite kernel as $k_{f}(\Tilde{x}_i, \Tilde{x}_j) = k_{\textrm{constant}} + k_{\textrm{matern}} + k_{\textrm{white}}$ where $k_{\textrm{constant}} = 1^2$, $k_{\textrm{white}} = \Lambda^2$, and $k_{\textrm{matern}}(\Tilde{x}_i, \Tilde{x}_j)$. 
$\Lambda$ is the noise level for the white kernel. We consider $l=2, \nu=1.5$, and $\Lambda=1$. 

\subsection{eXtreme Gradient Boosting (XGBoost)}
XGBoost~\cite{chen2016xgboost} is an algorithm that uses a decision tree as its base classifier for the target dataset $D_{1:t}^L$, that contains $L$ observations. In the typical Gradient Boosting (GB) algorithms, we use $B$ additive functions for $G$ times boosting of the gradient, to predict the output. Consider $f_k(\Tilde{x})$ as the prediction for the $k$-th instance at the $b$-th boost
\begin{equation}
    f_k(\Tilde{x}) \sim \sum_{b=1}^B f_b (\Tilde{x}_k).
\end{equation}
GB minimizes a loss function: 

\begin{equation}
O_b = \sum_{k=1}^L e(f_k(\Tilde{x}), \hat{f}_k(\Tilde{x})),
    \label{eq:GBloss}
\end{equation}

\noindent where $e(f_k(\Tilde{x}), \hat{f}_k(\Tilde{x}))$ is the measurement of the difference between the prediction~$f_k(\Tilde{x})$ and its real value~$\hat{f}_k(\Tilde{x})$.

If we add a regularization term $\Omega(f_b)$ to~\eqref{eq:GBloss}, the result will be the loss function of XGBoost:
\begin{align}
O_b & = \sum_{k=1}^L e(f_k(\Tilde{x}), \hat{f}_k(\Tilde{x})) + \sum_{b=1}^B \Omega(f_b) \nonumber\\
& = \sum_{k=1}^L e(f_k(\Tilde{x}), \hat{f}_k(\Tilde{x})) + \gamma \tau + 0.5 \lambda \lVert \omega \rVert^2.
    \label{eq:XGBloss}
\end{align}

The regularization term~$\Omega(f_b)$ penalizes the complexity of the model, and can be expressed as $\gamma \tau + 0.5 \lambda \lVert \omega \rVert^2$. Where $\tau$ represents the number of leaves in the tree filled with data, and $\gamma$ is the minimum loss reduction threshold for further partition. If the loss reduction is less than $\gamma$, XGBoost will stop. $\lambda$ is a fixed coefficient, and $\lVert \omega \rVert^2$ is the Euclidean norm of the leaf weight~\cite{wang2019xgboost}. 


\subsection{Multi Layer Perceptron (MLP)}

MLP~\cite{murtagh1991multilayer} is a class of artificial neural network that consists of multiple layers of neurons, each layer fully connected to the next one. Usually the structure includes an input layer, one or more hidden layers, and an output layer. In MLP, each hidden layer $g$ (where $1\leq g\leq G$) transforms the output of the previous layer $\Tilde{x}^{g-1}$ using a weight matrix $\varpi^g$ and a bias vector $\beta^g$. The transformation is a linear combination followed by a nonlinear activation $\sigma$:

\begin{equation}
    \hat{f}(\Tilde{x})^g = \sigma (\varpi^{g} \hat{f}(\Tilde{x})^{g-1} + \beta^g).
    \label{eq:MLP}
\end{equation}

\noindent In this paper, we used ReLU as the activation function $\sigma$. The final layer G produces the output. We also use L-BFGS~\cite{liu1989limited} to update the parameters using the gradient of the R-squared as the loss function.

 \subsection{Gaussian Process Regression (GPR)}

GPR~\cite{williams1995gaussian} is an algorithm that for each dimension $z =~1, \dots, L$ of the difference vector~$\Delta x_t$, estimates $f$ as
\begin{equation}
    f(\Tilde{x}) \sim \mathcal{GP}(\mu_{f}(\Tilde{x}), k_{f}(\Tilde{x}, \Tilde{x}')).
    \label{eq:GP}
\end{equation}
For the target dataset $D_{1:t}^L$, the trained GPR model can be queried at a new input point $\Tilde{x}_*$:
\begin{equation}
p(f(\Tilde{x}_*)|D_{1:t}^L, \Tilde{x}_*) = \mathcal{N}(\mu_{f}(\Tilde{x}_*), \sigma_{f}^2(\Tilde{x}_*)).
\end{equation}
Unlike other regression methods, GPR does not provide a prediction set. Instead, it provides two lists of means and variances for each prediction. The mean and variance predictions are calculated by a kernel vector~$\mathbf{k}_{f} = k(D_{1:t}^L, \Tilde{x}_*)$, and a kernel matrix~$K_{f}$, with entries of~$K_{f}^{ij} = k_{f}(\Tilde{x}_i, \Tilde{x}_j)$ as
\begin{align}
    \mu_{f}(\Tilde{x}_*) & = \mathbf{k}_{f}^\top K_{f}^{-1}D_{1:t}^L\nonumber \\
    \sigma_{f}^2(\Tilde{x}_*) & = k_{f}(\Tilde{x}_*, \Tilde{x}_*) - \mathbf{k}_{f}^\top K_{f}^{-1}\mathbf{k}_{f}.
    \label{eq:mean_variance}
\end{align}

\noindent where $k_f$ is the composite kernel used in SVR. To leverage the same evaluation used for other models, we incorporate the mean values of the GPR output.

\section{Experimental Setup}
In our experiments, we use the MyoPro, a lightweight two degrees of freedom upper limb exoskeleton~\cite{myomo-inc-cambridge-ma-no-date}. This wearable robot utilizes four surface electromyography (sEMG) sensors. Sensor placement includes two on the upper arm and two on the forearm. The device allows for user-specific threshold adjustments at these sensor locations, which differentiate between the muscular activities of the biceps and triceps for arm movements, and those related to the hand's opening and closing gestures. Activation of the device’s motor occurs upon exceeding these predefined thresholds, thereby facilitating user assistance. While the system provides data concerning the velocity of the integrated motors, it lacks the capability to offer information pertaining to the rotational movements of the hand. 

To collect data for modeling the robot, we add two APDM opal IMU sensors to the arm and wrist locations to measure the rotations of the hand (see Fig.~\ref{fig:exoimu}). Having access to the APDM opal IMU gyroscopes, we use Unscented Kalman filter~\cite{wan2000unscented} for calculation of the angles from angular velocity readings. 

We conducted data collection for six distinct tasks: Horizontal (H), Vertical (V), diagonal from Left leg to the Right eye (LR), diagonal from Right leg to the Left eye (RL), Eating (E), and Pushing (P) shown in Fig.~\ref{fig:tasks}. These tasks are chosen due to their diverse features. Horizontal~(H) involves a movement along the horizontal plane while Vertical (V) is a movement along the vertical plane. On the other hand, diagonal from Left leg to Right eye (LR) and diagonal from Right leg to Left eye (RL) are movements that cross the body. The last two tasks replicate the daily activities of the user where Eating (E) involves a range of motions towards the mouth, and Pushing (P) is a movement in the outward direction. 

For the horizontal task, participants moved an empty can between two predefined points on a table. The vertical task involved rotating the arm around the axis originating from the shoulder. We also introduced two diagonal tasks to complement horizontal and vertical tasks, requiring participants to move their hand from their leg to the front of their eye\textemdash either from the left leg towards the right eye or the opposite. The last two tasks are eating and pushing. In the eating task, participants moved their wrist from the table towards their mouth, while in the pushing task, they performed a forward arm movement, closing and opening the arm starting from the chest.


 \begin{figure}[h!]
    \centering
    \includegraphics[trim=0cm 0cm 0cm 0cm, clip, width=\columnwidth]{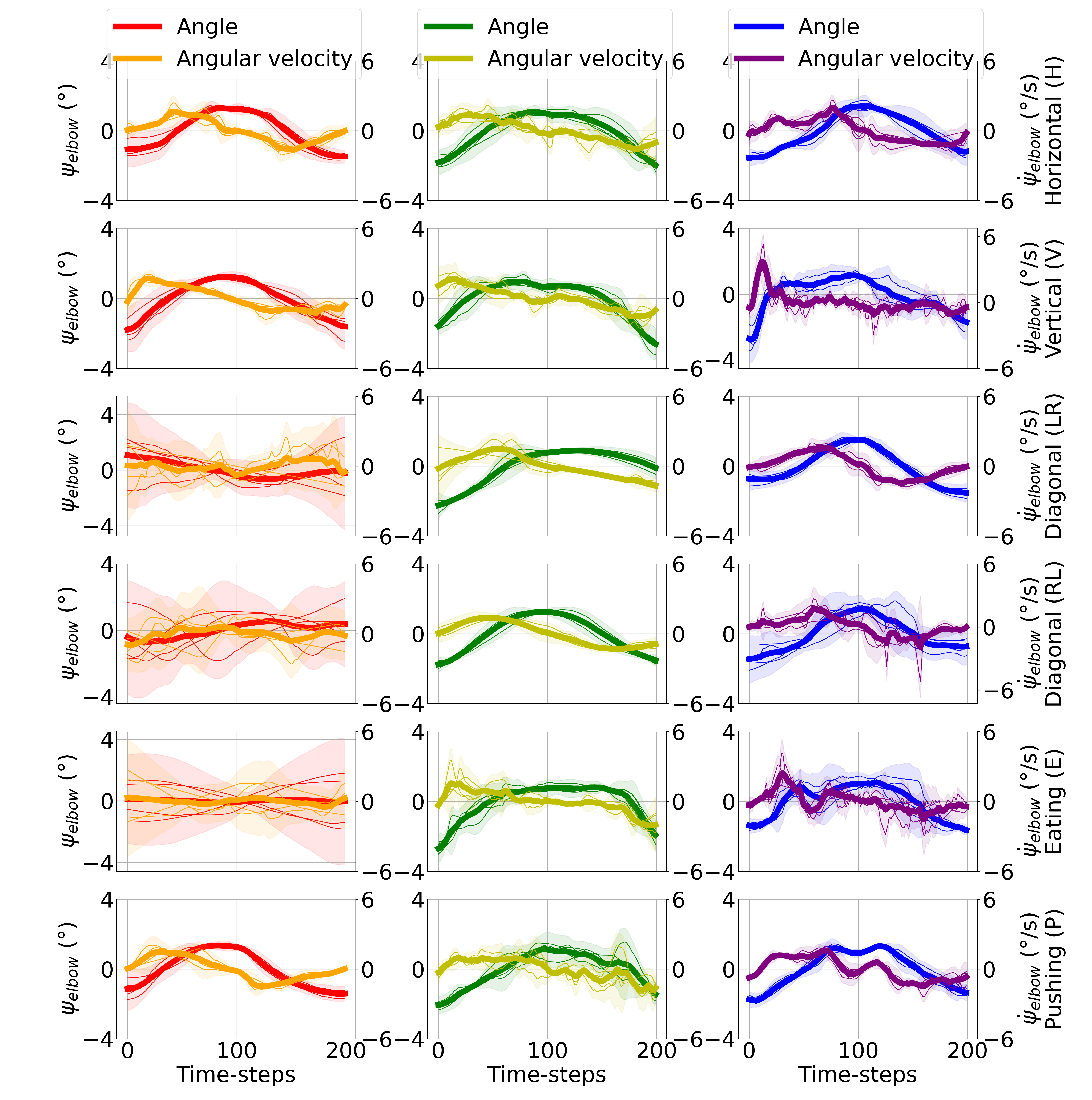}
    \caption{\small{Mean and standard deviation of the first two input features (elbow angle and angular velocity) calculated with $99\%$ confidence interval from four trials for each of the six tasks, with data collected from three test subjects distinguished by different colors from left to the right.}}
    \label{fig:angle}
\end{figure}

\section{Data Processing and Validation}

As illustrated in Fig.~\ref{fig:angle}, four separate trials for each task is separated from the data. We then averaged over the number of trials (four) to construct our dataset. To have a similar interval length across different tasks dataset, we employed one-dimensional linear interpolation. We linearly rescaled the inputs to have zero mean and unit variance on the training set. Although it is possible to similarly rescale the output data, we chose not to do this because our tests showed that it did not significantly enhance our results.

All of the six experiments were repeated by three users (among the authors), and the data was used for modeling by six different regression methods discussed in Sec~\ref{sec:Dynamic Model Learning through Regression}. 

We divided the dataset into five distinct subsets for cross-validation. The performance of the models were evaluated by averaging the outcomes of the R-squared score, expressed in percentage terms, across the cross-validation subsets. This metric was chosen over other evaluation criteria due to its universal applicability across all models and methods used in our study. 

\section{Results \& Discussion}



Our main goal is to determine the capability of a model, trained on a specific task (Fig.~\ref{fig:tasks}) to generalize to other tasks. We evaluate the generalizability of the obtained models when trained on a specific task and tested on other tasks. The complication is due to the fact that the trained model must be capable of approximating twelve different features as mentioned in Sec.~\ref{probForm}. For instance as it is shown in Fig.~\ref{fig:angle} while the elbow angle follows a roughly similar pattern in different tasks, the angular velocity varies significantly from one task to the other. 

\begin{figure}[h!]
    \centering
    \includegraphics[trim=0cm 0cm 0cm 0cm, clip, width=.99\columnwidth]{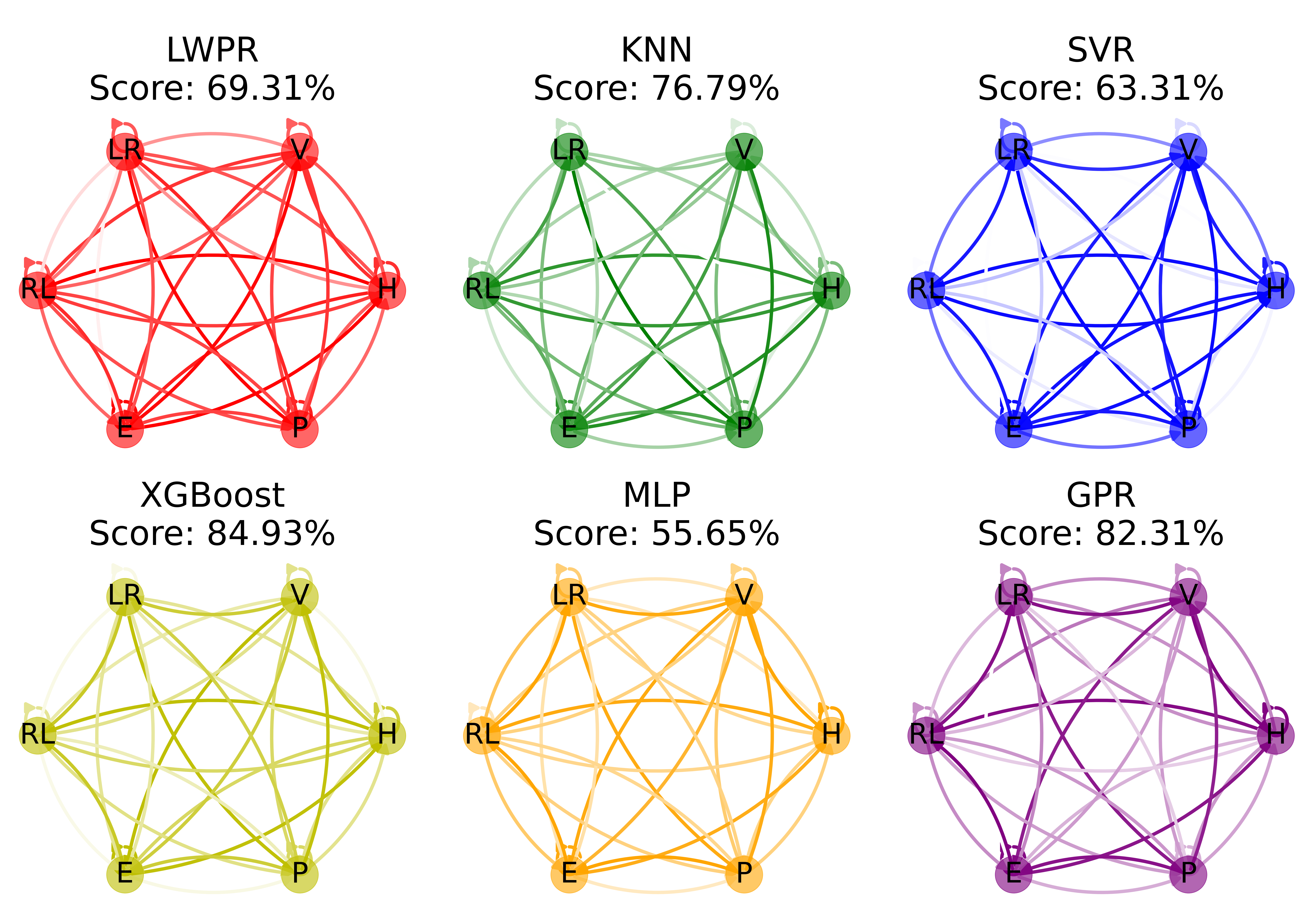}

    \caption{\small{Performance evaluation of each model, where training is conducted on the home node and testing is performed on the destination node data. Edge color intensity inversely correlates with the models' ability to generalize; a brighter edge shows lower generalizability.}}
    \label{fig:graphs}
\end{figure}

We use a graph representation to demonstrate the results. The nodes indicate tasks, while the edges show level of generalizability of the model when tested on the task in the destination node. As shown in Fig.~\ref{fig:graphs}, each model was trained on the home task, \textit{node} and tested on the destination task \textit{node}. The brightness of the \textit{edges} indicates the R-squared score for that specific training averaged over the models trained for three subjects. Thus a brighter or darker edge means lower or higher generalizability, respectively. It also allows us to assign a score based on the average number of input/output edge weights to the graphs, and order the models from the highest to the lowest. Based on their corresponding scores in Fig.~\ref{fig:graphs}, we notice that in terms of generalizability the selected regression algorithms can be sorted from best to worst as: XGBoost~(84.93\%), GPR~(82.31\%), KNN~(76.79\%), LWPR~(69.31\%), SVR~(63.31\%), and MLP~(55.65\%).

The difference in performances could have various reasons. For instance, the performance of LWPR is highly dependent on the choice of the hyper-parameter $r$, which defines the initial distance between local models. To have a fair comparison, we tuned $r$ once the model was being trained, and kept it fixed when testing on other task datasets. 

\begin{figure}[h!]
    \centering
    \includegraphics[width=\columnwidth]{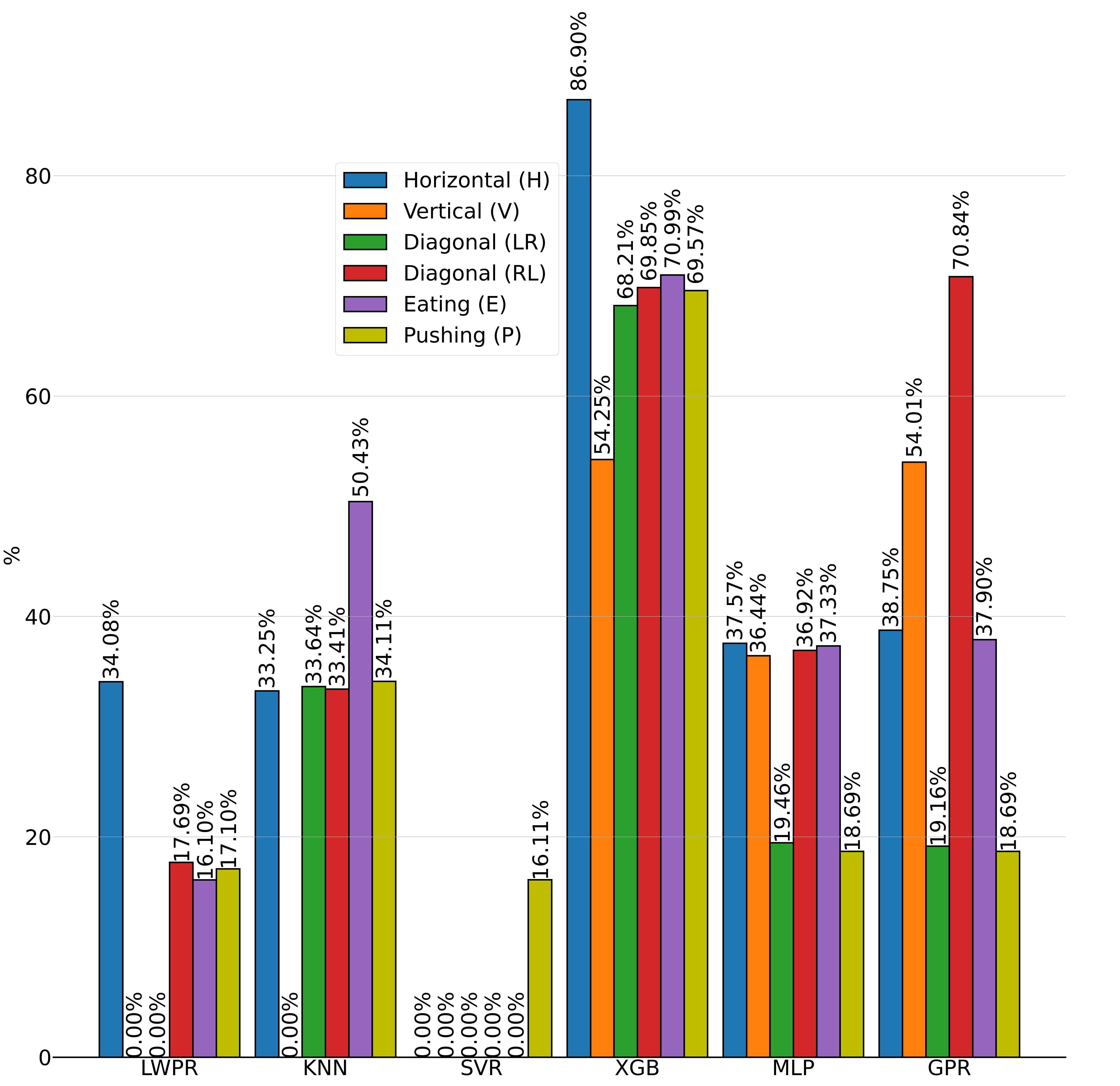}
    \caption{\small{The generalizability of different task data sets within each algorithm, using a R-squared value of 80\% as the threshold for acceptable performance.}}
    \label{fig:generalizability}
\end{figure}

In addition to our finding about the generalizability of the trained models, by considering an R-squared value of 80\% as the threshold for acceptable performance, and counting the output edges of each node, we can assess the extent to which each task model is generalizable within each algorithm. Fig.~\ref{fig:generalizability} shows the results for this evaluation. We notice that on a model trained by LWPR algorithm the H task model has the highest generalizability in comparison to the set of (RL, E, and P) task models that are 16.39\%, 17.98\%, and 16.98\% less generalizable respectively. In this algorithm the V and the LR task models do not generalize to any other model with an over 80\% R-squared. When using KNN, the E task model is the most generalizable and the set of (H, LR, RL, and P) are 17.18\%, 16.79\%, 17.02\%, and 16.32\% less generalizable and the V task is not generalizable over the threshold. In SVR only the model trained on the task P is 16.11\% generalizable and other task models generalize below the 80\% threshold. 

In the XGBoost that was ranked first in terms of generalizability, the H task on average generalizes by 86.90\% to the other tasks and V, LR, RL, E, and P task models are 32.65\%, 18.69\%, 17.05\%, 15.91\%, 17.33\% less generalizable than it. When ranking the models on their level of generalizability, we noticed that in MLP that is the least generalizable, all of the task models are nearly generalizable by 36~37\%. Except for the LR and the P that are 18\% less generalizable than the others. Our second place in the ranking list was for GPR that has a different generalizability from others. When using GPR, the RL is the most generalizable task model and V, H, E, LR, and P are respectively less generalizable models by 16.83\%, 32.09\%, 32.94\%, 51.68\%, 52.15\%.

\begin{figure}[ht]
    \centering
    \includegraphics[width=.9\columnwidth]{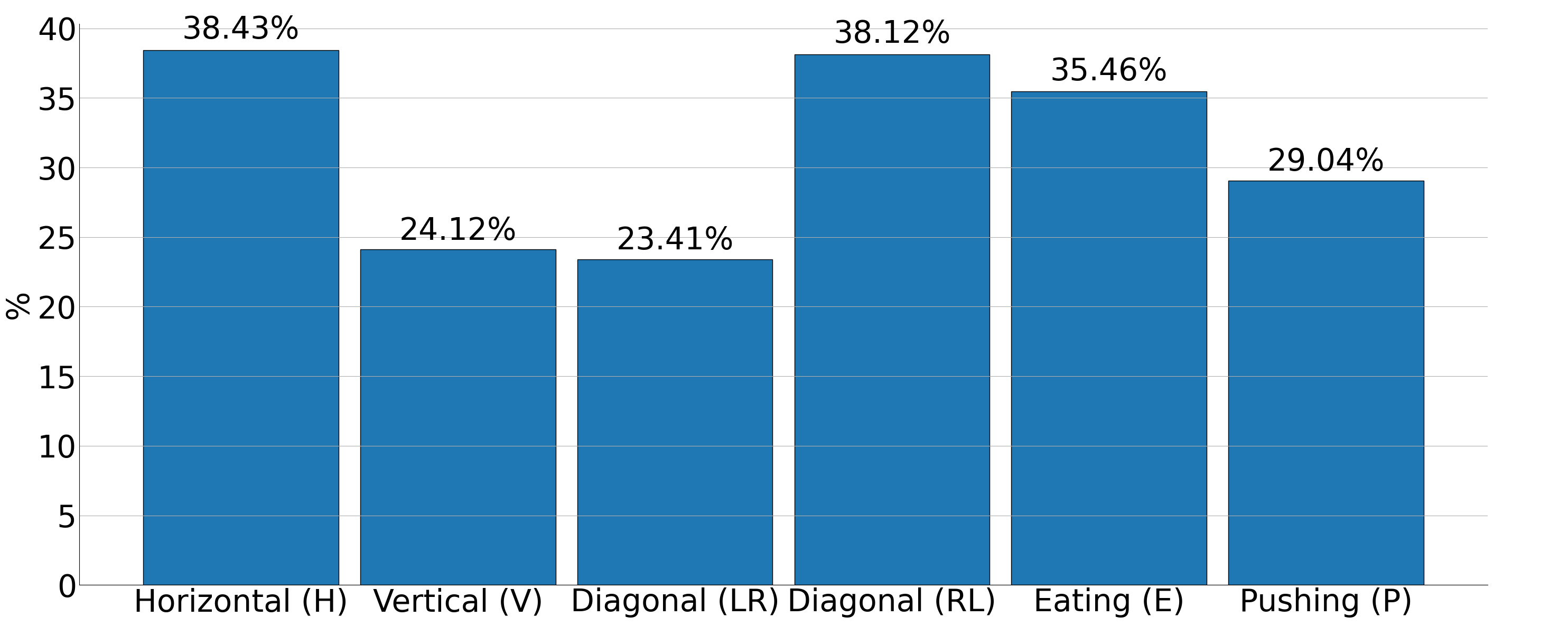}
    \caption{\small{The average generalizability for the task datasets in our study, using an R-squared value of 80 as the threshold for acceptable performance.}}
    \label{fig:taskGeneralizability}
\end{figure}

We then evaluated the algorithm-agnostic generalizability of the task models by summing their levels of generalizability in each algorithm and averaging over the total number of tasks. Fig.~\ref{fig:taskGeneralizability} shows that we can order the task models in our study for descending generalizability as H, RL, E, P, V, and LR.

The average training times for LWPR, KNN, SVR, XGBoost, MLP, and GPR in our study averaged over 5 folds of cross validation sets, six tasks and three subjects, were 0.003, 0.034, 0.005, 0.147, 0.247, and 0.243 seconds, respectively. This information can offer a more holistic view of the model performance when integrated with the generalizability.


\section{Conclusion}

In this paper, we focused on task generalizability in the data-driven modeling of assistive robots, particularly focusing on upper-limb exoskeletons. Our study employed six regression modeling techniques, among which the dynamic model constructed using the XGBoost algorithm showed superior generalizability capabilities. We collected the data from six different tasks, with an emphasis on cross-validating models trained on each one. The main finding from our experiments was the superiority of the model built for the H task in terms of generalizability. This insight provides context for refining the selection of model learning techniques and choosing appropriate tasks with specific features for model training. This study provides potential opportunities to construct more generalizable models that could lead to improved performance in a diverse array of tasks, ultimately benefiting the users of such assistive technologies.





 \section*{Acknowledgment}

This work was supported by the National Science Foundation (CMMI-2110214).

\bibliographystyle{IEEEtran}
\bibliography{bibtex/bib/root}

\section{Appendix}
\vspace{-10pt}

\begin{table*}
    	\centering
 \setlength\tabcolsep{1pt}
	\caption{\small{Performance Evaluation of the LWPR Model on Three Subjects. The table displays R-squared score, Root Mean Squared Error (RMSE), and Mean Absolute Error (MAE) metrics. Training datasets are represented vertically, and test datasets horizontally}}
	\label{table:LWPR}
\resizebox{.9\textwidth}{!}
{
\begin{tabular}{ c | c | c | c | c | c | c }

Subject 1 & \multicolumn{1}{c}{H} & \multicolumn{1}{c}{V} & \multicolumn{1}{c}{LR} & \multicolumn{1}{c}{RL} & \multicolumn{1}{c}{E} & \multicolumn{1}{c}{P}\\
&Total, $\psi_{elbow}$, $\Dot{\psi}_{wrist}$, $\psi_{elbow}$, $\Dot{\psi}_{wrist}$&Total, $\psi_{elbow}$, $\Dot{\psi}_{wrist}$, $\psi_{elbow}$, $\Dot{\psi}_{wrist}$&Total, $\psi_{elbow}$, $\Dot{\psi}_{wrist}$, $\psi_{elbow}$, $\Dot{\psi}_{wrist}$&Total, $\psi_{elbow}$, $\Dot{\psi}_{wrist}$, $\psi_{elbow}$, $\Dot{\psi}_{wrist}$&Total, $\psi_{elbow}$, $\Dot{\psi}_{wrist}$, $\psi_{elbow}$, $\Dot{\psi}_{wrist}$&Total, $\psi_{elbow}$, $\Dot{\psi}_{wrist}$, $\psi_{elbow}$, $\Dot{\psi}_{wrist}$\\
\hline
&\textbf{91.69, 97.02, 82.05, 97.86, 89.83}&92.92, 97.59, 84.28, 97.98, 91.84&92.53, 97.59, 88.35, 96.8, 87.39&94.88, 98.06, 90.53, 96.75, 94.16&83.02, 96.4, 73.95, 94.4, 67.32&83.6, 95.51, 82.92, 86.29, 69.67\\
H&\textbf{0.12, 0.03, 0.15, 0.04, 0.18}&0.13, 0.02, 0.14, 0.05, 0.2&0.09, 0.01, 0.09, 0.02, 0.16&0.08, 0.02, 0.09, 0.05, 0.13&0.08, 0.0, 0.05, 0.02, 0.15&0.18, 0.04, 0.22, 0.04, 0.27\\
LWPR (R=0.6)&\textbf{0.08, 0.02, 0.12, 0.03, 0.14}&0.07, 0.02, 0.1, 0.04, 0.15&0.05, 0.01, 0.07, 0.02, 0.11&0.05, 0.02, 0.07, 0.03, 0.09&0.04, 0.0, 0.04, 0.01, 0.11&0.1, 0.03, 0.13, 0.03, 0.2\\
\hline

&89.25, 95.06, 84.7, 95.76, 81.49&\textbf{88.11, 94.09, 75.83, 97.79, 84.72}&89.64, 96.33, 85.39, 93.9, 82.93&86.69, 95.69, 73.72, 96.83, 80.5&78.25, 93.35, 72.32, 86.33, 61.0&83.31, 96.31, 81.23, 76.19, 79.51\\
V&0.14, 0.03, 0.14, 0.06, 0.23&\textbf{0.14, 0.03, 0.15, 0.05, 0.23}&0.11, 0.02, 0.1, 0.03, 0.18&0.12, 0.02, 0.12, 0.04, 0.19&0.09, 0.0, 0.05, 0.02, 0.16&0.15, 0.04, 0.22, 0.04, 0.21\\
LWPR (R=0.8)&0.09, 0.03, 0.1, 0.04, 0.18&\textbf{0.08, 0.02, 0.1, 0.04, 0.16}&0.06, 0.01, 0.07, 0.02, 0.12&0.07, 0.01, 0.09, 0.03, 0.14&0.05, 0.0, 0.04, 0.02, 0.12&0.09, 0.03, 0.14, 0.03, 0.16\\
\end{tabular}
}
\end{table*}

\begin{table*}
    	\centering
 \setlength\tabcolsep{1pt}
	\caption{\small{Performance Evaluation of the KNN Model on Three Subjects. The table displays R-squared score, Root Mean Squared Error (RMSE), and Mean Absolute Error (MAE) metrics. Training datasets are represented vertically, and test datasets horizontally}}
	\label{table:KNN}
\resizebox{.9\textwidth}{!}
{
    \begin{tabular}{ c | c | c | c | c | c | c }

Subject 1 & \multicolumn{1}{c}{H} & \multicolumn{1}{c}{V} & \multicolumn{1}{c}{LR} & \multicolumn{1}{c}{RL} & \multicolumn{1}{c}{E} & \multicolumn{1}{c}{P}\\
&Total, $\psi_{elbow}$, $\Dot{\psi}_{wrist}$, $\psi_{elbow}$, $\Dot{\psi}_{wrist}$&Total, $\psi_{elbow}$, $\Dot{\psi}_{wrist}$, $\psi_{elbow}$, $\Dot{\psi}_{wrist}$&Total, $\psi_{elbow}$, $\Dot{\psi}_{wrist}$, $\psi_{elbow}$, $\Dot{\psi}_{wrist}$&Total, $\psi_{elbow}$, $\Dot{\psi}_{wrist}$, $\psi_{elbow}$, $\Dot{\psi}_{wrist}$&Total, $\psi_{elbow}$, $\Dot{\psi}_{wrist}$, $\psi_{elbow}$, $\Dot{\psi}_{wrist}$&Total, $\psi_{elbow}$, $\Dot{\psi}_{wrist}$, $\psi_{elbow}$, $\Dot{\psi}_{wrist}$\\
\hline

&\textbf{90.82, 96.48, 85.55, 95.98, 85.26}&85.53, 96.14, 60.49, 95.33, 90.14&89.84, 95.47, 84.55, 93.39, 85.95&92.47, 94.22, 87.88, 93.65, 94.12&81.68, 96.75, 69.11, 94.73, 66.1&84.54, 92.75, 85.36, 87.85, 72.21\\
H&\textbf{0.13, 0.03, 0.14, 0.06, 0.21}&0.16, 0.03, 0.23, 0.08, 0.22&0.1, 0.02, 0.11, 0.03, 0.16&0.09, 0.03, 0.11, 0.07, 0.13&0.08, 0.01, 0.06, 0.01, 0.15&0.17, 0.06, 0.2, 0.04, 0.26\\
KNN ($K_cv$=100)&\textbf{0.09, 0.03, 0.1, 0.05, 0.17}&0.1, 0.02, 0.14, 0.06, 0.17&0.06, 0.02, 0.08, 0.02, 0.12&0.06, 0.03, 0.08, 0.04, 0.1&0.04, 0.0, 0.04, 0.01, 0.11&0.1, 0.04, 0.14, 0.03, 0.2\\
\hline

&85.79, 94.85, 77.99, 95.72, 74.6&\textbf{89.31, 96.45, 75.63, 97.28, 87.88}&84.64, 95.99, 71.95, 93.59, 77.04&91.45, 95.97, 84.62, 96.39, 88.81&77.62, 95.65, 73.49, 92.23, 49.09&87.12, 95.33, 86.97, 87.14, 79.04\\
V&0.17, 0.04, 0.17, 0.06, 0.28&\textbf{0.16, 0.03, 0.18, 0.07, 0.24}&0.13, 0.02, 0.15, 0.03, 0.22&0.11, 0.03, 0.12, 0.05, 0.18&0.1, 0.01, 0.05, 0.02, 0.19&0.15, 0.04, 0.19, 0.04, 0.22\\
KNN ($K_cv$=100)&0.1, 0.03, 0.12, 0.05, 0.21&\textbf{0.1, 0.02, 0.12, 0.05, 0.2}&0.07, 0.01, 0.11, 0.02, 0.16&0.07, 0.02, 0.09, 0.04, 0.13&0.05, 0.0, 0.04, 0.01, 0.15&0.1, 0.04, 0.14, 0.03, 0.17\\
\end{tabular}
}
\end{table*}

\begin{table*}
    	\centering
 \setlength\tabcolsep{1pt}
	\caption{\small{Performance Evaluation of the SVR Model on Three Subjects. The table displays R-squared score, Root Mean Squared Error (RMSE), and Mean Absolute Error (MAE) metrics. Training datasets are represented vertically, and test datasets horizontally}}
	\label{table:SVR}
\resizebox{.9\textwidth}{!}
{
    \begin{tabular}{ c | c | c | c | c | c | c }

Subject 1 & \multicolumn{1}{c}{H} & \multicolumn{1}{c}{V} & \multicolumn{1}{c}{LR} & \multicolumn{1}{c}{RL} & \multicolumn{1}{c}{E} & \multicolumn{1}{c}{P}\\
&Total, $\psi_{elbow}$, $\Dot{\psi}_{wrist}$, $\psi_{elbow}$, $\Dot{\psi}_{wrist}$&Total, $\psi_{elbow}$, $\Dot{\psi}_{wrist}$, $\psi_{elbow}$, $\Dot{\psi}_{wrist}$&Total, $\psi_{elbow}$, $\Dot{\psi}_{wrist}$, $\psi_{elbow}$, $\Dot{\psi}_{wrist}$&Total, $\psi_{elbow}$, $\Dot{\psi}_{wrist}$, $\psi_{elbow}$, $\Dot{\psi}_{wrist}$&Total, $\psi_{elbow}$, $\Dot{\psi}_{wrist}$, $\psi_{elbow}$, $\Dot{\psi}_{wrist}$&Total, $\psi_{elbow}$, $\Dot{\psi}_{wrist}$, $\psi_{elbow}$, $\Dot{\psi}_{wrist}$\\
\hline

&\textbf{90.47, 82.74, 91.4, 93.41, 94.35}&85.4, 69.11, 83.91, 94.4, 94.15&70.06, 40.68, 88.31, 61.31, 89.93&86.43, 69.83, 90.66, 88.46, 96.78&36.66, -5.31, 48.58, 24.15, 79.23&80.21, 84.71, 93.7, 53.56, 88.87\\
H&\textbf{0.1, 0.07, 0.1, 0.08, 0.13}&0.13, 0.08, 0.14, 0.09, 0.17&0.1, 0.08, 0.09, 0.08, 0.14&0.09, 0.08, 0.09, 0.09, 0.09&0.08, 0.03, 0.07, 0.06, 0.11&0.12, 0.08, 0.13, 0.08, 0.16\\
SVR&\textbf{0.08, 0.07, 0.09, 0.07, 0.11}&0.1, 0.07, 0.11, 0.08, 0.13&0.08, 0.07, 0.08, 0.07, 0.11&0.08, 0.07, 0.08, 0.09, 0.08&0.06, 0.03, 0.06, 0.05, 0.09&0.09, 0.07, 0.1, 0.07, 0.12\\
\hline

&90.18, 85.49, 90.63, 94.65, 89.94&\textbf{87.46, 69.6, 89.73, 95.33, 95.2}&69.91, 45.18, 84.53, 62.71, 87.24&87.39, 73.91, 89.82, 89.81, 96.03&35.99, -5.31, 50.14, 25.87, 73.28&82.03, 86.51, 94.64, 54.87, 92.09\\
V&0.11, 0.07, 0.11, 0.07, 0.17&\textbf{0.11, 0.08, 0.12, 0.08, 0.15}&0.11, 0.07, 0.1, 0.08, 0.16&0.09, 0.07, 0.09, 0.09, 0.1&0.08, 0.03, 0.07, 0.06, 0.13&0.11, 0.08, 0.12, 0.08, 0.14\\
SVR&0.08, 0.06, 0.09, 0.06, 0.13&\textbf{0.09, 0.08, 0.09, 0.07, 0.12}&0.08, 0.07, 0.09, 0.07, 0.11&0.08, 0.06, 0.08, 0.08, 0.09&0.06, 0.03, 0.06, 0.05, 0.11&0.09, 0.07, 0.1, 0.07, 0.11\\
\end{tabular}
}
\end{table*}

\begin{table*}
    	\centering
 \setlength\tabcolsep{1pt}
	\caption{\small{Performance Evaluation of the XGBoost Model on Three Subjects. The table displays R-squared score, Root Mean Squared Error (RMSE), and Mean Absolute Error (MAE) metrics. Training datasets are represented vertically, and test datasets horizontally}}
	\label{table:XGBoost}
\resizebox{.9\textwidth}{!}
{
    \begin{tabular}{ c | c | c | c | c | c | c }

Subject 1 & \multicolumn{1}{c}{H} & \multicolumn{1}{c}{V} & \multicolumn{1}{c}{LR} & \multicolumn{1}{c}{RL} & \multicolumn{1}{c}{E} & \multicolumn{1}{c}{P}\\
&Total, $\psi_{elbow}$, $\Dot{\psi}_{wrist}$, $\psi_{elbow}$, $\Dot{\psi}_{wrist}$&Total, $\psi_{elbow}$, $\Dot{\psi}_{wrist}$, $\psi_{elbow}$, $\Dot{\psi}_{wrist}$&Total, $\psi_{elbow}$, $\Dot{\psi}_{wrist}$, $\psi_{elbow}$, $\Dot{\psi}_{wrist}$&Total, $\psi_{elbow}$, $\Dot{\psi}_{wrist}$, $\psi_{elbow}$, $\Dot{\psi}_{wrist}$&Total, $\psi_{elbow}$, $\Dot{\psi}_{wrist}$, $\psi_{elbow}$, $\Dot{\psi}_{wrist}$&Total, $\psi_{elbow}$, $\Dot{\psi}_{wrist}$, $\psi_{elbow}$, $\Dot{\psi}_{wrist}$\\
\hline

&\textbf{97.73, 99.47, 95.58, 99.63, 96.23}&96.83, 99.12, 92.56, 99.63, 96.0&97.01, 99.31, 96.27, 99.05, 93.41&98.61, 99.52, 96.98, 99.52, 98.42&93.42, 98.54, 92.05, 97.83, 85.26&97.1, 99.44, 96.78, 97.68, 94.5\\
H&\textbf{0.07, 0.01, 0.07, 0.02, 0.11}&0.09, 0.01, 0.1, 0.02, 0.14&0.06, 0.01, 0.05, 0.01, 0.11&0.04, 0.01, 0.05, 0.02, 0.06&0.05, 0.0, 0.03, 0.01, 0.1&0.08, 0.02, 0.1, 0.02, 0.11\\
XGBoost&\textbf{0.04, 0.01, 0.05, 0.01, 0.08}&0.05, 0.01, 0.07, 0.02, 0.09&0.03, 0.01, 0.04, 0.01, 0.06&0.03, 0.01, 0.04, 0.01, 0.04&0.02, 0.0, 0.02, 0.01, 0.06&0.04, 0.01, 0.07, 0.01, 0.08\\
\hline

&96.92, 99.33, 93.33, 99.5, 95.53&\textbf{96.95, 99.22, 92.39, 99.71, 96.46}&96.19, 98.68, 93.43, 98.41, 94.23&98.38, 99.49, 96.26, 99.56, 98.19&92.91, 98.28, 89.43, 98.29, 85.65&96.89, 99.47, 96.65, 96.51, 94.94\\
V&0.08, 0.02, 0.09, 0.02, 0.12&\textbf{0.08, 0.01, 0.09, 0.02, 0.13}&0.07, 0.01, 0.07, 0.01, 0.11&0.05, 0.01, 0.06, 0.02, 0.07&0.05, 0.0, 0.03, 0.01, 0.09&0.08, 0.02, 0.1, 0.02, 0.11\\
XGBoost&0.05, 0.01, 0.07, 0.02, 0.09&\textbf{0.04, 0.01, 0.07, 0.02, 0.09}&0.03, 0.01, 0.04, 0.01, 0.06&0.03, 0.01, 0.04, 0.01, 0.05&0.02, 0.0, 0.02, 0.01, 0.06&0.04, 0.01, 0.07, 0.01, 0.08\\
\end{tabular}
}
\end{table*}

\begin{table*}
    	\centering
 \setlength\tabcolsep{1pt}
	\caption{\small{Performance Evaluation of the MLP Model on Three Subjects. The table displays R-squared score, Root Mean Squared Error (RMSE), and Mean Absolute Error (MAE) metrics. Training datasets are represented vertically, and test datasets horizontally}}
	\label{table:MLP}
\resizebox{.9\textwidth}{!}
{
    \begin{tabular}{ c | c | c | c | c | c | c }

    Subject 1 & \multicolumn{1}{c}{H} & \multicolumn{1}{c}{V} & \multicolumn{1}{c}{LR} & \multicolumn{1}{c}{RL} & \multicolumn{1}{c}{E} & \multicolumn{1}{c}{P}\\
    &Total, $\psi_{elbow}$, $\Dot{\psi}_{wrist}$, $\psi_{elbow}$, $\Dot{\psi}_{wrist}$&Total, $\psi_{elbow}$, $\Dot{\psi}_{wrist}$, $\psi_{elbow}$, $\Dot{\psi}_{wrist}$&Total, $\psi_{elbow}$, $\Dot{\psi}_{wrist}$, $\psi_{elbow}$, $\Dot{\psi}_{wrist}$&Total, $\psi_{elbow}$, $\Dot{\psi}_{wrist}$, $\psi_{elbow}$, $\Dot{\psi}_{wrist}$&Total, $\psi_{elbow}$, $\Dot{\psi}_{wrist}$, $\psi_{elbow}$, $\Dot{\psi}_{wrist}$&Total, $\psi_{elbow}$, $\Dot{\psi}_{wrist}$, $\psi_{elbow}$, $\Dot{\psi}_{wrist}$\\
\hline

&\textbf{98.85, 99.77, 98.12, 99.91, 97.61}&97.69, 99.37, 93.52, 99.86, 98.03&99.16, 99.51, 98.02, 99.68, 99.43&99.48, 99.66, 98.78, 99.79, 99.68&92.36, 83.84, 92.99, 99.19, 93.42&98.81, 99.67, 98.17, 99.22, 98.17\\
H&\textbf{0.05, 0.01, 0.05, 0.01, 0.08}&0.06, 0.01, 0.07, 0.01, 0.08&0.02, 0.01, 0.03, 0.01, 0.03&0.02, 0.01, 0.03, 0.01, 0.03&0.03, 0.01, 0.02, 0.01, 0.06&0.05, 0.01, 0.07, 0.01, 0.06\\
MLP&\textbf{0.03, 0.0, 0.04, 0.01, 0.06}&0.03, 0.01, 0.05, 0.01, 0.06&0.01, 0.0, 0.02, 0.01, 0.02&0.01, 0.01, 0.02, 0.01, 0.02&0.02, 0.0, 0.02, 0.0, 0.04&0.03, 0.01, 0.05, 0.01, 0.04\\
\hline

&97.84, 99.67, 96.0, 99.71, 95.99&\textbf{98.39, 99.72, 96.27, 99.87, 97.71}&98.22, 99.19, 95.49, 99.59, 98.59&99.26, 99.79, 97.99, 99.91, 99.35&93.71, 95.73, 92.46, 97.48, 89.18&98.08, 99.58, 97.06, 98.69, 97.02\\
V&0.06, 0.01, 0.07, 0.02, 0.09&\textbf{0.06, 0.01, 0.06, 0.02, 0.09}&0.03, 0.01, 0.05, 0.01, 0.05&0.03, 0.0, 0.04, 0.01, 0.04&0.04, 0.01, 0.02, 0.01, 0.08&0.06, 0.01, 0.08, 0.01, 0.07\\
MLP&0.03, 0.01, 0.05, 0.01, 0.07&\textbf{0.03, 0.01, 0.05, 0.01, 0.07}&0.02, 0.01, 0.03, 0.01, 0.04&0.02, 0.0, 0.02, 0.01, 0.03&0.02, 0.0, 0.02, 0.01, 0.06&0.03, 0.01, 0.05, 0.01, 0.05\\
\end{tabular}
}
\end{table*}

\begin{table*}
    	\centering
 \setlength\tabcolsep{1pt}
	\caption{\small{Performance Evaluation of the GPR Model on Three Subjects. The table displays R-squared score, Root Mean Squared Error (RMSE), and Mean Absolute Error (MAE) metrics. Training datasets are represented vertically, and test datasets horizontally}}
	\label{table:GP}
\resizebox{.9\textwidth}{!}
{
    \begin{tabular}{ c | c | c | c | c | c | c }
    
         Subject 1 & \multicolumn{1}{c}{H} & \multicolumn{1}{c}{V} & \multicolumn{1}{c}{LR} & \multicolumn{1}{c}{RL} & \multicolumn{1}{c}{E} & \multicolumn{1}{c}{P}\\
         &Total, $\psi_{elbow}$, $\Dot{\psi}_{wrist}$, $\psi_{elbow}$, $\Dot{\psi}_{wrist}$&Total, $\psi_{elbow}$, $\Dot{\psi}_{wrist}$, $\psi_{elbow}$, $\Dot{\psi}_{wrist}$&Total, $\psi_{elbow}$, $\Dot{\psi}_{wrist}$, $\psi_{elbow}$, $\Dot{\psi}_{wrist}$&Total, $\psi_{elbow}$, $\Dot{\psi}_{wrist}$, $\psi_{elbow}$, $\Dot{\psi}_{wrist}$&Total, $\psi_{elbow}$, $\Dot{\psi}_{wrist}$, $\psi_{elbow}$, $\Dot{\psi}_{wrist}$&Total, $\psi_{elbow}$, $\Dot{\psi}_{wrist}$, $\psi_{elbow}$, $\Dot{\psi}_{wrist}$\\
\hline

&\textbf{98.1, 99.9, 96.13, 99.88, 96.51}&96.52, 99.32, 90.64, 99.74, 96.38&97.96, 99.74, 95.98, 99.49, 96.64&99.13, 99.8, 97.93, 99.7, 99.08&94.46, 99.31, 92.29, 98.99, 87.25&96.83, 99.45, 95.72, 98.72, 93.43\\
H&\textbf{0.06, 0.01, 0.07, 0.01, 0.1}&0.09, 0.01, 0.11, 0.02, 0.13&0.05, 0.01, 0.05, 0.01, 0.08&0.03, 0.01, 0.04, 0.01, 0.05&0.05, 0.0, 0.03, 0.01, 0.09&0.09, 0.02, 0.11, 0.01, 0.12\\
GPR&\textbf{0.03, 0.0, 0.05, 0.01, 0.08}&0.04, 0.01, 0.07, 0.01, 0.09&0.02, 0.0, 0.03, 0.01, 0.05&0.02, 0.0, 0.03, 0.01, 0.03&0.02, 0.0, 0.02, 0.0, 0.05&0.04, 0.01, 0.07, 0.01, 0.08\\
\hline

&97.45, 99.77, 94.74, 99.72, 95.57&\textbf{97.7, 99.76, 94.1, 99.92, 97.02}&96.44, 99.68, 93.51, 99.41, 93.17&99.02, 99.84, 97.21, 99.93, 99.08&92.68, 98.96, 91.44, 97.57, 82.77&97.94, 99.8, 96.85, 99.12, 95.99\\
V&0.07, 0.01, 0.08, 0.02, 0.11&\textbf{0.07, 0.01, 0.09, 0.01, 0.12}&0.07, 0.01, 0.07, 0.01, 0.11&0.04, 0.01, 0.05, 0.01, 0.05&0.05, 0.0, 0.03, 0.01, 0.1&0.07, 0.01, 0.09, 0.01, 0.1\\
GPR&0.04, 0.01, 0.05, 0.01, 0.08&\textbf{0.04, 0.01, 0.06, 0.01, 0.08}&0.02, 0.0, 0.04, 0.01, 0.06&0.02, 0.0, 0.04, 0.0, 0.04&0.02, 0.0, 0.02, 0.01, 0.07&0.04, 0.01, 0.06, 0.01, 0.07\\
\end{tabular}
}
\end{table*}

\addtolength{\textheight}{-12cm}   





\end{document}